\documentclass[11pt]{article}
\usepackage[margin=1in]{geometry}
\usepackage{amsmath, amssymb}
\usepackage{graphicx}
\usepackage{braket}
\usepackage{cite}
\usepackage{hyperref}
\hypersetup{
    colorlinks=true,
    linkcolor=blue,
    citecolor=blue,
    urlcolor=blue
}

\title{Rethinking LLM Training through Information Geometry and Quantum Metrics}
\author{Riccardo Di Sipio \\ \small Dayforce, HCM \\ \small \texttt{riccardo.disipio@dayforce.com}}
\date{\today}

\begin{document}

\maketitle

\begin{abstract}
Optimization in large language models (LLMs) unfolds over high-dimensional parameter spaces with non-Euclidean structure. Information geometry frames this landscape using the Fisher information metric, enabling more principled learning via natural gradient descent \cite{amari1998natural, amari2016information}. Though often impractical \cite{martens2015optimizing}, this geometric lens clarifies phenomena such as sharp minima, generalization, and observed scaling laws \cite{kaplan2020scaling}. We argue that curvature-aware approaches deepen our understanding of LLM training. Finally, we speculate on quantum analogies based on the Fubini–Study metric and Quantum Fisher Information \cite{bengtsson2017geometry, petz2011introduction}, hinting at efficient optimization in quantum-enhanced systems.
\end{abstract}

\section{Introduction}

The optimization of large language models (LLMs) has revealed both remarkable successes and profound theoretical puzzles. As these models scale, they exhibit smoother loss landscapes, better generalization, and empirically predictable performance. These trends are formalized into \emph{scaling laws} which relate computation, data, and parameter count to loss~\cite{kaplan2020scaling,hoffmann2022training}. However, the reasons behind these patterns remain incompletely understood. What governs the shape of the loss landscape at scale? Why do some architectures converge more efficiently than others? And are our optimization tools even suited to the true geometry of the learning process?

\paragraph{}
In this paper, we explore a speculative but structured hypothesis: that certain characteristics of the LLM training dynamics, particularly those that involve curvature, convergence, and generalization, may be more clearly understood through the lens of \emph{quantum geometry}. Although quantum mechanics and deep learning operate in different domains, both describe high-dimensional systems that evolve according to variational principles and exhibit sensitivity to local curvature. Quantum systems evolve in manifolds endowed with \emph{Fubini--Study metric}, a Riemannian geometry that defines the distances between pure quantum states. This metric induces \emph{quantum Fisher information matrix} (QFI), which measures local sensitivity to parameter changes and encodes a sharper, more expressive geometry than its classical counterpart~\cite{helstrom1976quantum,liu2020quantum}.

\paragraph{}
In contrast, classical deep learning uses curvature-based methods such as \emph{Fisher information matrix} only when explicitly approximated, as in natural gradient descent~\cite{amari2016information}. But these methods are rarely practical at scale: computing or inverting the Fisher matrix is computationally expensive and unstable in high dimensions. Quantum systems, however, \emph{implement their optimization geometry intrinsically}. They do not approximate the curvature; they embed it. This asymmetry highlights a key difference: classical models struggle to access second-order structure, whereas quantum systems inherently optimize on a richly curved manifold.

From this perspective, training an LLM is not unlike a quantum system collapsing to a measurement outcome. Gradient descent acts as a noisy, iterative projection onto a lower-loss state, shaped by stochastic data exposure and local curvature. Analogies to wavefunction collapse, energy minimization, and superposition offer a new vocabulary for describing how large models move through parameter space and, perhaps, new tools for improving their training.

\paragraph{}
The remainder of this paper develops this argument in detail. Section~\ref{sec:background} introduces the relevant concepts in optimization, Fisher geometry, and quantum state space. Section~\ref{sec:related} situates our work in the broader landscape of classical and quantum machine learning literature. Section~\ref{sec:discussion} discusses implications for optimization, scaling behavior,  algorithm design, and outlines future directions. We conclude by reflecting on the value of metaphor as a tool for bridging disciplines, and on the possibility that quantum geometry may teach us to train better models.

\section{Theoretical Background}
\label{sec:background}
This section introduces the key ideas behind the optimization of LLMs and the foundational geometric structures of quantum mechanics. While these domains differ in application and context, they share a surprising amount of mathematical scaffolding. Our aim here is not to formalize every element, but to build intuition for the analogies explored in later sections.

\subsection{Optimization and Geometry in LLMs}

Modern large language models (LLMs) rely on first-order optimization techniques such as stochastic gradient descent (SGD) and its variants \cite{kingma2014adam, bottou2018optimization}. While these methods are effective in high-dimensional settings, they operate on parameter spaces that may not be well-represented by Euclidean geometry. Instead, tools from \textit{information geometry} provide a richer picture of the underlying optimization landscape \cite{amari1998natural, amari2016information}.

This behavior invites a geometric interpretation: training a neural network can be viewed as traversing a high-dimensional manifold shaped by the model’s parameters and the loss function. The Fisher information matrix plays a central role in understanding this geometry. It defines a local Riemann metric over parameter space, capturing the curvature of the loss surface and the sensitivity of the model’s output distribution to small changes in its parameters.
While the Fisher information is often introduced from a probabilistic or statistical standpoint, its geometric interpretation (the metric tensor on the statistical manifold) is what links it directly to variational principles in physics. This connection sets the stage for the analogies to come.

This parallel extends even further. In general relativity, gravity is not a force in the traditional sense but the visible effect of spacetime curvature—objects move as they do because of the geometry they inhabit. In deep learning, especially in LLM, a similar effect is observed in word embedding geometry: semantically related words appear to 'attract' each other in vector space. Yet this attraction is not fundamental but the consequence of a deeper structure: the high-dimensional probability distribution modeled by the neural network. The embedding space reflects the curvature of this distribution, just as gravitational trajectories reflect the curvature of spacetime.

\subsection{Geometry of Probability Spaces and Fisher Information}

Given a parametric probability distribution $p(x; \boldsymbol{\theta})$, the space of parameters $\boldsymbol{\theta}$ forms a statistical manifold. To understand curvature on this space, we begin with the Euclidean distance:
\[
d(x, y) = \sqrt{(x^i - y^i)(x^j - y^j)}
\]

In contrast, the geometry of a curved statistical manifold is defined by a metric tensor:
\[
g_{ij}(\boldsymbol{\theta}) = \mathbb{E} \left[ \frac{\partial \log p(x; \boldsymbol{\theta})}{\partial \theta^i} \frac{\partial \log p(x; \boldsymbol{\theta})}{\partial \theta^j} \right]
\]
This is the Fisher information matrix, and it captures the local curvature by measuring how distinguishable two infinitesimally close distributions are.

For a parametrized curve $\theta(t)$ on the manifold, the squared infinitesimal length is given by:
\[
ds^2 = g_{ij}(\boldsymbol{\theta})\, d\theta^i d\theta^j
\]

This construction is closely related to Riemannian geometry. Just as general relativity describes gravity as curvature induced by energy and mass in spacetime, information geometry describes the learning landscape as curvature induced by the informational structure of the model. In this analogy, the Fisher information matrix plays a role similar to the metric tensor in general relativity: it determines how distances are measured, how geodesics are computed, and ultimately, how optimization proceeds on the manifold.

This parallel extends even further. In general relativity, gravity is not a force in the traditional sense but the visible effect of spacetime curvature in which objects move as they do because of the geometry they inhabit. In deep learning, especially in LLM, a similar effect can be observed in the geometry of word embeddings: semantically related words appear to 'attract' each other in vector space. Yet this attraction is not fundamental but is the consequence of a deeper structure: the high-dimensional probability distribution modeled by the neural network. The embedding space reflects the curvature of this distribution, just as gravitational trajectories reflect the curvature of spacetime.

\begin{figure}[h]
  \centering
  \includegraphics[width=0.6\linewidth]{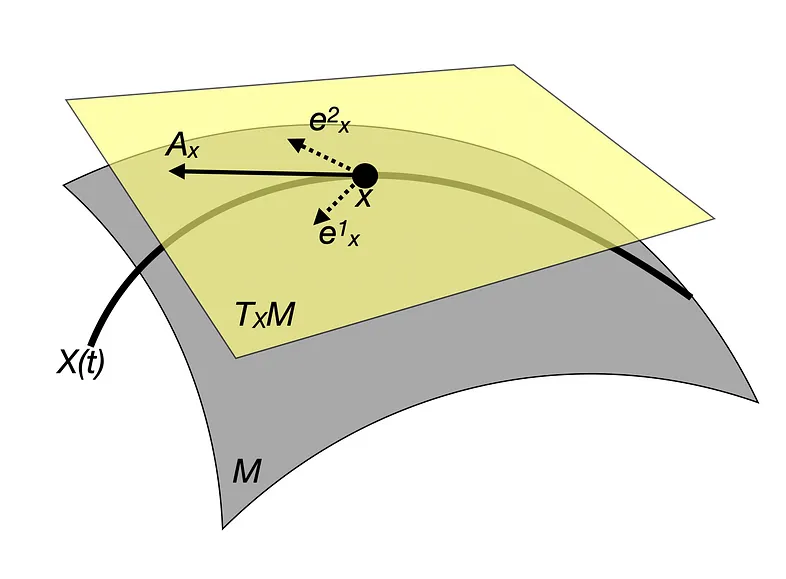}
  \caption{Curve on a manifold illustrating tangent space and information geometry concepts. A curve $X(t)$ on manifold $M$ passes through point $x$. The tangent space at $x$, denoted as $T_xM$, is depicted as a plane. The tangent vector $\mathbf{A}_x$ is shown, expressed in basis $\{\mathbf{e}^1_x,
\mathbf{e}^2_x\}$. Tangent vectors derived from the log-
likelihood function $\log p(x; \boldsymbol{\theta})$ define the local structure of information geometry.}
  \label{fig:tangent_space}
\end{figure}

As shown in Figure \ref{fig:tangent_space}, the local geometry of a statistical manifold can be characterized using a tangent space at each point, where tangent vectors correspond to infinitesimal changes in parameters. The metric tensor (in our case derived from the Fisher Information) defines inner products in this space, enabling the computation of distances and gradients.

\subsection{Gradient-Based Optimization and Curvature Awareness}

In Euclidean space, gradient descent follows the negative gradient of a scalar loss $L(\boldsymbol{\theta})$:
\[
\theta^{(t+1)} = \theta^{(t)} - \eta \, \nabla L
\]

However, on a curved manifold, the gradient must account for the metric. The gradient of $L$ becomes:
\[
\nabla L = g^{ij} \frac{\partial L}{\partial \theta^j}
\]

This leads to \emph{natural gradient descent}, which rescales the update direction using the inverse of the Fisher information:
\[
\theta^{(t+1)} = \theta^{(t)} - \eta \, F^{-1} \nabla_\theta L
\]

While this method is theoretically appealing, computing and inverting $F$ is often impractical for large-scale models, This is a limitation we revisit later.

\subsection{A Glimpse into Quantum Geometry}
Quantum mechanics describes the evolution of physical systems not by point particles, but by state vectors in complex Hilbert spaces. Unlike classical systems, where probabilities are assigned to definite outcomes, quantum systems exist in superpositions, and their dynamics are governed by linear, unitary transformations. The geometry of this space is not Euclidean, but projective: global phases are physically irrelevant, so the true configuration space is the set of rays in Hilbert space.
This leads naturally to the Fubini-Study metric, a Riemannian metric defined on the space of pure quantum states. It measures the distinguishability of nearby quantum states, taking into account the complex structure and phase invariance. Crucially, it is this metric that induces the quantum Fisher information matrix (QFI), the quantum counterpart to the classical Fisher matrix [3]. The QFI governs how sensitively a quantum system responds to changes in its parameters, which is precisely the type of curvature measure that also shapes learning dynamics in deep networks.
Whereas the classical Fisher information captures sensitivity of likelihood functions to parameter variation, the QFI captures how quickly a quantum state changes under infinitesimal parameter shifts, embedding a richer geometric structure. In this sense, the quantum manifold is “sharper”: its local geometry is more curved, suggesting that optimization over such a manifold may follow steeper, more directed gradients.

\subsection{Scaling Laws and the Limits of Classical Geometry}
One of the most intriguing discoveries in deep learning over the past several years is the emergence of scaling laws: predictable relationships between model size, dataset size, compute budget, and performance [4,5]. These laws show that, up to certain limits, larger models trained with more compute on more data achieve systematically lower loss. However, they also exhibit diminishing returns, with each doubling of compute yielding smaller performance gains.
This invites speculation: are these limits fundamental, or merely artifacts of classical architectures and training methods? If the geometry induced by classical Fisher information places implicit constraints on the paths optimization can follow, perhaps a richer geometry (like that of quantum manifolds) could break through these scaling ceilings. Quantum-inspired training could, in principle, allow for more efficient exploration of parameter space, greater separation between local minima, and more sharply curved descent pathways.

\section{Related Work and Theoretical Context}\label{sec:related}
Understanding the training dynamics of large neural networks has been a subject of intense interest across multiple disciplines, ranging from statistical physics and differential geometry to information theory and complexity science. Similarly, the exploration of quantum machine learning (QML) has grown rapidly, with efforts to either simulate quantum behavior in classical models or accelerate learning using quantum hardware. In this section, we briefly survey the major threads that intersect with our discussion.

\subsection{Geometry and Curvature in Deep Learning}
Several foundational works have explored the geometric properties of deep neural networks. Amari’s work on information geometry \cite{amari1998natural} introduced the Fisher information matrix as a natural Riemannian metric on parameter space, providing a basis for natural gradient descent, an optimization method that accounts for the manifold structure of statistical models. More recently, Martens \cite{martens2015optimizing} and others have revisited this idea in the context of scalable deep learning, showing how curvature-based methods can improve convergence in high-dimensional, ill-conditioned loss landscapes.
Pennington and Bahri \cite{pennington2017geometry} have contributed significantly to the understanding of loss surfaces in wide neural networks using tools from random matrix theory. Their work reveals that curvature (as captured by the Hessian or Fisher matrix) not only guides optimization, but also relates closely to generalization and mode connectivity.
These studies support the intuition that the geometry of the optimization manifold matters deeply, and that Fisher information is a key player in shaping it.

\subsection{Quantum Information Geometry and the Quantum Fisher Information} 

In quantum theory, the quantum Fisher information matrix (QFI) plays an analogous but richer role. It quantifies the distinguishability of nearby quantum states, and governs bounds on parameter estimation via the quantum Cramér–Rao inequality \cite{petz2011introduction, bengtsson2017geometry}. In the quantum setting, the Cramér–Rao bound generalizes through the Quantum Fisher Information, providing a lower bound on the variance of unbiased estimators.
Unlike its classical counterpart, the QFI arises from the Fubini-Study metric on the projective Hilbert space, which reflects the true geometric structure of quantum states.

In quantum information geometry, the Fubini–Study metric defines a Riemannian structure on the projective Hilbert space of pure states. Given a normalized quantum state $\ket{\psi(\theta)}$ parameterized by $\theta$, the Fubini–Study line element is:

\[
ds^2 = \braket{d\psi | d\psi} - |\braket{\psi | d\psi}|^2
\]

This metric captures how distinguishable nearby quantum states are and plays a similar role to the Fisher Information metric in classical information geometry.

The Quantum Fisher Information (QFI) extends this idea to mixed states represented by a density matrix $\rho(\theta)$. The QFI matrix is defined as:

\[
F_{ij} = \mathrm{Tr}\left[\rho(\theta) \, L_i L_j\right]
\]

where $L_i$ is the symmetric logarithmic derivative (SLD), implicitly defined by:

\[
\frac{\partial \rho(\theta)}{\partial \theta^i} = \frac{1}{2} \left( \rho L_i + L_i \rho \right)
\]

This formulation naturally generalizes the classical Fisher metric, and underlies the geometry of quantum statistical models. In our speculative analogy, we suggest that systems which optimize according to the Fubini–Study geometry may have access to more efficient paths through parameter space, effectively embodying a form of natural gradient descent without approximation.

The QFI has been used to study quantum phase transitions, entanglement structure, and variational quantum algorithms. These are domains where sensitivity to parameter changes plays a critical role \cite{petz2011introduction,bengtsson2017geometry}. Recent works have explored the implications of QFI in optimization contexts, suggesting that its sharper curvature may aid convergence in variational quantum circuits (VQCs) \cite{schuld2019evaluating, wierichs2021avoiding}.
These insights lend credence to the idea that curvature-rich metrics can improve optimization, and provide theoretical grounding for the analogies drawn in this paper.

\subsection{Quantum-Inspired and Hybrid Models}
A growing body of research investigates the use of quantum-inspired classical models or hybrid quantum-classical architectures. For example, quantum kernels \cite{havlivcek2019supervised} and tensor network models \cite{stoudenmire2016supervised} have been proposed as ways to simulate quantum entanglement on classical hardware. Others have examined the information content of parameterized quantum circuits and their expressivity relative to deep neural networks \cite{sim2019expressibility}.
More broadly, the field of quantum machine learning (QML) has explored whether quantum systems can offer computational advantages for training, representation, or inference. While practical quantum advantage remains elusive, the field continues to evolve, and several benchmarks have demonstrated that quantum models may represent certain functions more compactly or learn them with fewer queries under ideal conditions.
These developments support the claim that quantum geometry is not merely metaphorical, but has the potential to reshape the conceptual foundations of learning theory, even in classical contexts.

\section{Discussion and Future Directions}\label{sec:discussion}
The analogies explored in this paper are not merely aesthetic; they suggest practical and theoretical implications for how we think about deep learning optimization, model scaling, and geometry-based learning. While the connection between quantum mechanics and machine learning remains largely conceptual at this stage, it opens several promising avenues for future exploration.

\subsection{Reconsidering the Role of Geometry in Optimization}
The geometry induced by the classical Fisher information matrix has long been recognized as a key to understanding the behavior of the model. Algorithms such as natural gradient descent leverage this insight to follow geodesics on the parameter manifold, adapting optimization steps to local curvature. However, in practice, these methods are rarely used at scale: computing or inverting the Fisher matrix is prohibitively expensive for modern architectures.
By contrast, quantum systems inherently evolve over curved manifolds whose geometry is governed by the Fubini-Study metric. This metric defines the shortest paths between quantum states in the projective Hilbert space, and its associated quantum Fisher information matrix (QFI) emerges naturally from the structure of the system. In this sense, quantum systems "bake in" the geometry of optimization, representing local curvature with little or no approximation, simply as a consequence of their evolution.
This contrast is striking: while classical deep learning must estimate or approximate information geometry through costly computations, quantum systems implement it intrinsically. The implication is that optimization on a quantum manifold may inherently follow more informed, curvature-aware trajectories, without requiring additional algorithmic machinery.
This raises the intriguing possibility that what we observe as semantic similarity in embedding space is analogous to gravity in physics: not a primitive force, but a visible consequence of underlying curvature. In this view, neural networks do not 'learn meaning' but rather induce geometry in parameter space; the effects we associate with intelligence or coherence emerge as geodesic alignments within this curved manifold.

\subsection{Implications for Scaling Laws}
As discussed in Section \ref{sec:background}, current scaling laws reveal diminishing returns for increasing model size or compute. This raises the question: are these power-law trends fundamental to the classical geometry of model space, or might they be altered under different assumptions?
The quantum geometric perspective opens the door to alternative scaling regimes. If QFI-like curvature permits more efficient discrimination of function classes, or enables faster convergence by sharpening gradient trajectories, then it is conceivable that quantum-enhanced models may deviate from classical scaling trends, especially in data-sparse or noise-sensitive regimes.
Testing this hypothesis would require either simulation of quantum optimization geometries or experimentation with hybrid models that approximate QFI behavior, both tractable research directions.

\subsection{Inspirations for Algorithm Design}
A practical takeaway from this observation is that optimization algorithms inspired by quantum geometry may offer structural advantages. In classical settings, curvature-aware methods based on the Fisher matrix are often abandoned due to computational cost. But if quantum systems naturally encode the Fubini-Study geometry, effectively embedding second-order information into their evolution, then training quantum models is equivalent to performing information-geometric optimization by default.
This insight invites the development of quantum-inspired approximations in classical models: architectures or parameterizations that mimic the behavior of systems evolving under the Fubini-Study metric. Alternatively, hybrid quantum-classical systems could be designed to let quantum components handle the geometrically complex parts of the optimization landscape, while classical components perform update rules or evaluations.
In both cases, the core idea remains: where classical models struggle to approximate curvature, quantum models live in it.

\subsection{Limits and Cautions}
Of course, there are limits to this analogy. Neural networks are not quantum systems. Their parameters evolve under different physical laws, and while the mathematics of quantum theory may offer powerful conceptual tools, they must be applied with care. The goal here is not to claim equivalence, but to propose a productive metaphor, a lens that may sharpen our understanding of systems that are otherwise difficult to analyze.
Additionally, the practical challenges of quantum machine learning such as hardware fragility, limited qubit counts, and barren plateaus, remain unresolved. Any benefit from quantum geometry must be weighed against these constraints.

\subsection{Future Work}
Several research directions emerge from this perspective:
Formal derivations comparing the geometry of classical and quantum loss landscapes.
Simulation of QFI-based optimization on classical models.
Visualization tools for understanding curvature differences in parameter space.
Exploration of hybrid LLMs with quantum-inspired modules or initialization schemes.
Empirical tests of scaling behavior under QFI-motivated training schemes.
By treating LLM optimization as a form of quantum-like collapse over curved information manifolds, we gain new language and tools to reason about generalization, curvature, and convergence. Whether or not this path leads to better models, it offers a uniquely interdisciplinary lens, one that blends machine learning with some of the most successful abstractions in the history of physics.

\section{Conclusions}
In this paper, we explored conceptual analogies between the training dynamics of large language models (LLMs) and the geometry of quantum systems. By drawing connections between gradient descent and wavefunction collapse, loss landscapes and energy potentials, and classical versus quantum Fisher information, we proposed a perspective in which deep learning optimization is viewed as a trajectory through a high-dimensional information manifold.
Central to this view is the role of geometry. Classical learning algorithms can, in theory, benefit from curvature-aware methods such as those based on the Fisher information matrix. However, these approaches are rarely used at scale due to their computational cost and instability. Quantum systems, by contrast, evolve according to the Fubini-Study metric, a natural Riemannian geometry that directly induces the quantum Fisher information matrix. As a result, quantum systems intrinsically encode second-order information, allowing them to represent and traverse curved optimization landscapes with little or no approximation.
This distinction suggests a possible path forward: quantum-enhanced or quantum-inspired systems may offer a fundamentally different way to learn, not by adding complexity to approximate curvature, but by existing in a geometry where curvature is already present. Whether this leads to better generalization, faster convergence, or new scaling behaviors remains an open question. But the analogy is not just aesthetic; it invites us to rethink how we understand learning itself.

\bibliographystyle{plain}
\bibliography{bibliography}

\end{document}